\documentclass[a4paper, 10pt, conference]{ieeeconf}      
\usepackage{FG2024}

\FGfinalcopy 

\IEEEoverridecommandlockouts                              
\usepackage{graphics} 
\usepackage{epsfig} 
\usepackage{mathptmx} 
\usepackage{times} 
\usepackage{amsmath} 
\usepackage{amssymb}  
\usepackage{multirow}  

\title{\LARGE \bf
PhySU-Net: Long Temporal Context Transformer for rPPG with Self-Supervised Pre-training
}

\author{\parbox{16cm}{\centering
    {\large Marko Savic$^1$ and Guoying Zhao$^2$}\\
    {\normalsize
     Center for Machine Vision and Signal Analysis (CMVS),
 University of Oulu,
 FI-90014, Finland}}
}

\begin{document}

\ifFGfinal
\thispagestyle{empty}
\pagestyle{empty}
\else
\author{Anonymous FG2024 submission\\ Paper ID \FGPaperID \\}
\pagestyle{plain}
\fi
\maketitle

\begin{abstract}

Remote photoplethysmography (rPPG) is a promising technology that consists of contactless measuring of cardiac activity from facial videos. Most recent approaches utilize convolutional networks with limited temporal modeling capability or ignore long temporal context. Supervised rPPG methods are also severely limited by scarce data availability. In this work, we propose PhySU-Net, the first long spatial-temporal map rPPG transformer network and a self-supervised pre-training strategy that exploits unlabeled data to improve our model. Our strategy leverages traditional methods and image masking to provide pseudo-labels for self-supervised pre-training. Our model is tested on two public datasets (OBF and VIPL-HR) and shows superior performance in supervised training. Furthermore, we demonstrate that our self-supervised pre-training strategy further improves our model's performance by leveraging representations learned from unlabeled data.

\end{abstract}


 \section{Introduction}
\label{sec:intro}

Physiological signals like the blood volume pulse (BVP) are used to determine vital healthcare parameters such as heart rate (HR), heart rate variability (HRV), respiratory frequency (RF)~\cite{breathe} and oxygen saturation (SpO2). Moreover, they are psychological indicators since they change accordingly with emotional states~\cite{multiemotion}. A convenient non-contact method, which employs cheap and ubiquitous RGB cameras, is remote photoplethysmography (rPPG) ~\cite{verkruysse_remote_2008}. Similar to contact PPG, rPPG relies on capturing periodical variations in optical absorption of tissue caused by cardiac activity. The main differences between rPPG and contact PPG are in optical sensor (RGB camera vs. photodiode), distance from sensor (meters vs. millimeters) and lighting source (LED vs complex environmental). With the weak signal derived from skin color variation, a camera also captures overwhelming environmental noise caused by lighting changes, subject movement and sensor variations. Accurately and robustly identifying the faint quasi-periodical rPPG signal is a challenging task. Several early methods had been proposed that relied on optical/physiological considerations expressed through mathematical models like CHROM~\cite{de_haan_robust_2013}, POS~\cite{wang_algorithmic_2016}, PBV~\cite{de_haan_improved_2014}, LGI~\cite{pilz_local_2018}  or common blind source separation approaches such as ICA~\cite{poh_advancements_2010} and PCA~\cite{lewandowska_measuring_2011}. However, since they lacked robustness in scenarios with variable light and movement, they were surpassed by deep learning approaches.
Most deep learning methods employed Convolutional Neural Networks (CNN). Early 2D-CNN models extracted HR from adjacent frames, such as HR-CNN~\cite{spetlik_visual_2018} and DeepPhys~\cite{chen_deepphys_2018}. End-to-end 3D-CNN models such as PhysNet~\cite{yu2019remote}, rPPGNet~\cite{yu2019remotecomp} and AutoHR~\cite{autohr} exploited the temporal information. Non-end-to-end models used spatial-temporal maps, less affected by noise, like RhythmNet~\cite{niu_rhythmnet_2020}, CVD~\cite{niu_video-based_2020}, Dual-GAN~\cite{lu_dual-gan_2021} and BVPNet~\cite{das_bvpnet_nodate}. Recently, self-attention based transformer architectures 
have been proposed for rPPG, like EfficientPhys~\cite{liu_efficientphys_2021}, Physformer~\cite{yu_physformer_2022}, TransPPG~\cite{kang_transppg_2022}, RADIANT~\cite{gupta2023radiant}, TransPhys~\cite{wang2023transphys}. TransRPPG~\cite{yu2021transrppg}, also deals with rPPG, but with face presentation attack detection as a novel application. However, none of the aforementioned works fully exploit the temporal modeling capabilities of the transformer architecture by utilizing a long temporal context.  

Lack of labeled data is an issue in rPPG, as data collection is costly, requires medical devices and presents privacy concerns. Supervised methods struggle to achieve robustness and high generalization capability when trained on small datasets with specific noise distributions. To mitigate data scarcity in rPPG, natural images~\cite{niu_rhythmnet_2020}, synthetic signals~\cite{niu_synrhythm_2018} and synthetic avatar facial videos~\cite{mcduff_advancing_2020} have been used to learn generalized representations. Nonetheless, non-facial videos and synthetic data cannot replicate signals and environmental noise present in real facial video data. Several unsupervised contrastive methods have also been proposed to learn from data without labels~\cite{wang_self-supervised_2021,gideon_way_2021,sun_contrast-phys_2022}, but failed to reach close to supervised performance or to demonstrate their transferable capability.

The rPPG signal is quasi-periodical and drowned in environmental noise, but over a longer time frame it retains similar features (amplitude, frequency, dicrotic notch, systolic point, diastolic point). Consequently, modeling a longer time frame can aid in distinguishing the rPPG signal from noise. 
We propose a method for rPPG that capitalizes on the long-range capabilities and linear complexity of Swin transformer architecture~\cite{liu_swin_2021}, and makes use of compact noise-robust spatial-temporal maps~\cite{niu_video-based_2020}. Our model can learn rich features from a long input of \(\sim\)20s, making it more robust to environmental noise. We frame the signal prediction as an image (multi-signal) reconstruction task~\cite{das_bvpnet_nodate}, providing stronger supervision and enabling our model to train from scratch without any complex training strategies or data augmentation.
Moreover, we propose a self-supervised framework that allows our model to seamlessly learn rich representations by pre-training on unlabeled data. We leverage pseudo-labels that are generated via masking the input signals, enabling the model to learn temporal and frequency characteristics of rPPG signals from the input data alone, and use a traditional method to guide the model via an additional regression constraint. We clarify that our self-supervised pre-training is meant for subsequent fine-tuning, we do not attempt to surpass purely unsupervised methods on unlabeled data, but to produce useful representations for improving the downstream supervised learning task. 

Our contributions are as follows:
\begin{itemize}
    \item We propose PhySU-Net, the first long spatial-temporal map transformer network for rPPG, and show state-of-the-art performance on both OBF~\cite{li_obf_2018} and VIPL-HR~\cite{niu_vipl-hr_2018} datasets.
    \item We propose the first image-based pretext task self-supervised method for rPPG .
    \item We leverage traditional rPPG methods for a regression constraint that improves self-supervised learning.
\end{itemize}

\section{Methodology}
\begin{figure*}[!htbp]
\centerline{\includegraphics[width=\textwidth]{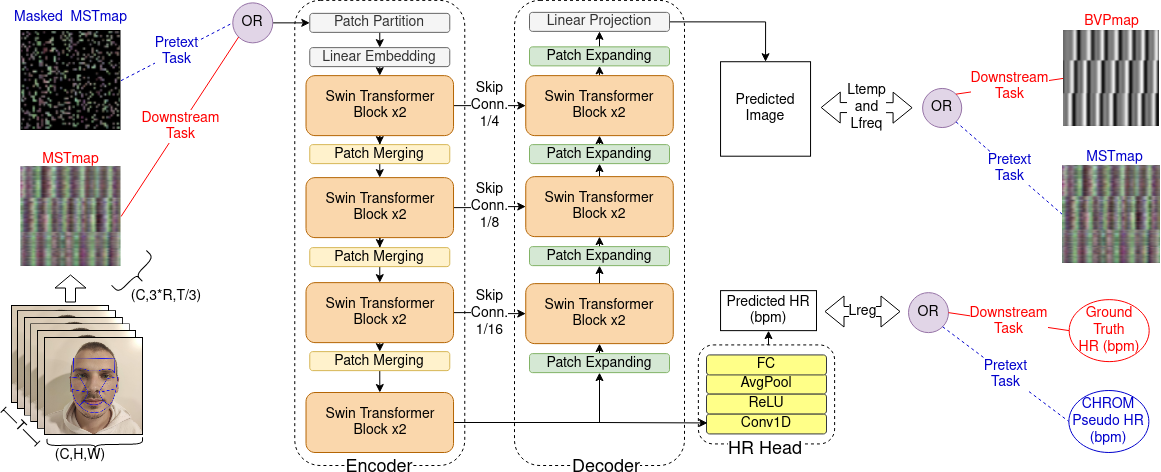}}
\caption{\textbf{Overview of our method:} The input video is processed into a stacked MSTmap. For the supervised downstream task, the decoder reconstructs an image with similar temporal and frequency properties as the BVPmap label, and the HR head regresses the HR value with the HR ground truth as label. For the self-supervised pretext task, only the input and the labels change. The input is a masked version of the MSTmap, that the decoder attempts to reconstruct into a full MSTmap. For the HR regression, a CHROM pseudo-label is used.}
\label{fig:main}
\end{figure*}

To address the issue of non-robustness to environmental noise in rPPG, we propose PhySU-Net, a long temporal context transformer network with spatial-temporal map input. To mitigate data scarcity, a self-supervised pre-training scheme is exploited in our network that can learn useful representations from unlabeled data. Our method is divided into preprocessing, PhySU-Net model and self-supervised pre-training. The complete overview is shown in Fig.~\ref{fig:main}.

\subsection{Preprocessing}

\begin{figure}[!htbp]
\centering
\includegraphics[width=0.5\textwidth]{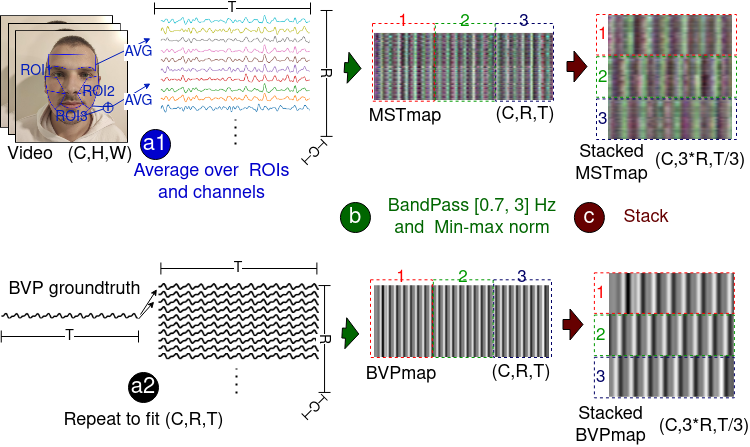}
\caption{\textbf{Preprocessing:} a1) \(C \times R\) temporal sequences are extracted by averaging pixels for each channel and ROI combination. a2) The ground truth BVP is duplicated to fit the dimension of the MSTmap b) The sequences are filtered with a pass band of \([0.7, 3] Hz\) and min-max normalized. c) The MSTmap and BVPmap is temporally stacked to form square images. }
\label{fig:prepro}
\end{figure}

Multi-scale spatial temporal maps (MSTmaps)~\cite{niu_video-based_2020} are a good intermediate representation of the rPPG signal, they are calculated by averaging pixels from regions of interest (ROI) on the face, that has been proven to successfully extract relevant features for rPPG as it is used by most traditional methods. MSTmaps are less affected by HR irrelevant noise than unprocessed input videos and are very compact, allowing us to model longer temporal sequences. We follow ~\cite{niu_video-based_2020} for the MSTmap generation procedure. For the image reconstruction target, BVPmaps are generated by stacking copies of the BVP ground truth, so that each row of the MSTmap will reconstruct the same BVP ground truth. Finally, both MSTmaps and BVPmaps are divided into three equal chunks along the temporal axis that are stacked to form a square image.
The preprocessing is shown in Fig.~\ref{fig:prepro}.

\subsection{PhySU-Net model}

We formulate a multitask learning problem comprised of HR regression and image-based rPPG signal prediction, proposing a revised Swin-Unet~\cite{cao_swin-unet_2021} to solve it. Swin~\cite{liu_swin_2021} transformer builds feature maps hierarchically by merging image patches and computes self-attention within local windows, allowing it to have linear complexity. It was adapted into a Swin-Unet~\cite{cao_swin-unet_2021} by constructing an encoder and decoder with skip connections and adding a new patch expanding layer for up-sampling. This architecture is particularly suited to the rPPG problem, as the self-attention mechanism allows for stronger modeling of long temporal sequences.
The rPPG signal reconstruction is framed as an MSTmap to BVPmap image reconstruction task~\cite{das_bvpnet_nodate}. Therefore, each coarse temporal input sequence is reconstructed into a clean rPPG signal, providing strong supervision and enabling our model to train easily from scratch.
We expand the Swin-Unet model to be better suited for our task with the following two additions. 
Firstly, the input MSTmap is stacked to a square size as shown in Fig.~\ref{fig:prepro} (c). This strategy allows us to maintain the temporal consistency in the patch sampling, and to inject temporally distant patches in the local neighborhoods of windowed self-attention, encouraging global level interaction in earlier layers.
Secondly, we add an HR regression head that is constructed of a 1d convolution, which aids faster convergence, followed by a ReLU non-linearity, adaptive average pooling and finally a fully connected layer predicting the HR value. The full architecture is shown in Fig.~\ref{fig:main}.

For HR regression an L1 loss is employed, and for the BVPmap reconstruction we utilize both a temporal and a frequency based loss. The temporal loss promotes high correlation between prediction and label, instead of the commonly used Pearson loss~\cite{yu2019remote} that is sensitive to synchronization errors, we utilize the maximum cross-correlation (\(MCC\))~\cite{gideon_way_2021}, which determines the correlation at an ideal offset and is invariant to phase differences between prediction and label. 
\begin{equation}
\small   MCC(x,y) = C_{pr}\times Max(\frac{F^{-1}\{BPass(F\{x\} \cdot \overline{ F\{ y\}}) \}}{\sigma_{x}\times\sigma_{y}}) 
   \label{eq:mcc}
\end{equation}
The \(MCC\), in Eq.~\ref{eq:mcc}, is the cross-correlation in the frequency domain scaled by \( C_{pr} \), the ratio of power inside the HR relevant frequencies.
Our temporal loss \( L_{temp} \), in  Eq.~\ref{eq:ltemp}, is the mean of the negative \(MCC\) of each of the \(C\times R\) temporal sequences from the prediction X and label Y.
\begin{equation}
\small
   L_{temp}(X,Y) = 1 - \frac{\sum\limits_{c=1}^{C} \sum\limits_{r=1}^{R} MCC( X(c,r,t),Y(c,r,t))}{C \times R}
      \label{eq:ltemp}
\end{equation}
Our frequency loss \( L_{freq} \), as shown in Eq.~\ref{eq:lfreq}, is defined as the mean squared error between the Power Spectral Densities (PSD) of the prediction and label temporal sequences. The error is squared to accentuate the peaks in the spectrum, as they are of most relevance.
\begin{equation}
\small   L_{freq}(X,Y) = \frac{\sum\limits_{c=1}^{C} \sum\limits_{r=1}^{R} (PSD(X(c,r,t))- PSD(Y(c,r,t))^{2}}{C \times R}
         \label{eq:lfreq}
\end{equation}
The final loss, in Eq.~\ref{eq:finalloss}, is the weighted sum of the regression, temporal and frequency losses.
\begin{equation}
\small   L = \alpha L_{reg} + \beta L_{temp} + \gamma L_{freq}
            \label{eq:finalloss}
\end{equation}

\subsection{Self-supervised pre-training}
We propose a self-supervised pre-training framework that enables our network to learn useful representations on unlabeled data. Our method consists of pseudo-HR regression and masked MSTmap reconstruction pretext tasks, that utilize the same losses and network as the downstream task, the only difference being the input and labels.
Inspired by Masked Auto Encoder~\cite{he_masked_2021}, we input a masked version of the MSTmap having the model reconstruct the missing parts, making the network learn temporal and frequency characteristics (by minimizing \(L_{temp}\) and \(L_{freq}\)) of the coarse rPPG signals present in the MSTmap. We mask 75\% of all 4x4 input patches for the image based task. For the HR regression task, we generate a pseudo-label with the traditional method CHROM~\cite{de_haan_robust_2013}, which serves as an additional constraint to guide the models' self-supervised learning. 
Our framework can be easily adapted to other image-based pretext tasks and pseudo-labels.

\section{Experiments}
We evaluate PhySU-Net on the OBF~\cite{li_obf_2018} and VIPL-HR~\cite{niu_vipl-hr_2018} datasets, and show superior performance compared to state-of-the-art methods. To prove the effectiveness of our self-supervised approach, a protocol similar to~\cite{wang_self-supervised_2021} is used, where linear classification (re-train only last fully-connected layer) and transfer learning (re-train whole network) are performed. Additionally, we provide ablation studies to analyze crucial network components, demonstrate the effectiveness of longer temporal context and show generalizability of our self-supervised pre-training method.
\subsection{Experimental Setup}
\textbf{Datasets}: OBF~\cite{li_obf_2018} contains 200 five-minute-long constant frame rate RGB videos recorded from 100 subjects, captured in a controlled environment with stable lighting and minimal movement of the subjects. 
VIPL-HR~\cite{niu_vipl-hr_2018} contains 2,378 RGB videos of 20s-30s length with variable and unstable frame rate. It was recorded in a challenging environment with different devices, large movements and various lighting. It contains many sources of noise, making HR estimation challenging.

\noindent \textbf{Evaluation Metrics}: 
We follow previous works by using absolute error (MAE), root-mean-square error (RMSE), standard deviation (SD) and Pearson's correlation coefficient (R). 

\noindent \textbf{Implementation}:
We choose T=576 (19.2s at 30fps) due to VIPL-HR~\cite{niu_vipl-hr_2018} videos being 20-30s long and for computational ease. For fair comparison with other methods, a five-fold subject exclusive cross validation is adopted for VIPL-HR and ten-fold for OBF in all experiments.
In training, the AdamW optimizer is used with epsilon=1e-8, betas=(0.9, 0.99), lr=5e-5, wd=0.05, batch=8. Loss parameters are set at \(\alpha=5\), \(\beta=1\), \(\gamma=5\). For supervised experiments and self-supervised pre-training 50 epochs are used, for all fine-tuning (linear and transfer) on VIPL-HR the epochs are lowered to 25. No data augmentation is used, training samples are 576 frames long with a fixed sliding window of 30 frames, testing samples are 576 frames long with no overlap.

\subsection{Experimental results}
\begin{table}[t!]
\begin{center}
\caption{Supervised training results on OBF~\cite{li_obf_2018} and VIPL-HR~\cite{niu_vipl-hr_2018}: divided in traditional, CNN and transformer based methods. Best results are marked in bold, second best in underline.}

\begin{tabular}{llll}   
\hline
OBF~\cite{li_obf_2018}         & \begin{tabular}[c]{@{}l@{}}RMSE \(\downarrow\)\\ \end{tabular} & \begin{tabular}[c]{@{}l@{}}SD \(\downarrow\)\\ \end{tabular} & r \(\uparrow\) \\ \hline
CHROM~\cite{de_haan_robust_2013}                                                  & 2.733    & 2.730    & 0.980 \\
POS~\cite{wang_algorithmic_2016}             & 1.906    & 1.899   & 0.991   \\
\hline
rPPGNet~\cite{yu2019remotecomp}            & 1.800      & 1.756   & 0.992    \\
CVD~\cite{niu_video-based_2020}                & 1.260     & 1.257   & 0.996   \\
\hline
Physformer~\cite{yu_physformer_2022}       & \underline{0.804}    & \underline{0.804}   & \underline{0.998}       \\
\textbf{PhySU-Net (Ours) } &   \textbf{0.659}       &  \textbf{0.618}       &     \textbf{0.999}       \\
\hline
\end{tabular}
\newline
\vspace{4pt}
\newline
\begin{tabular}{lllll}   
\hline
VIPL-HR~\cite{niu_vipl-hr_2018}          & \begin{tabular}[c]{@{}l@{}}MAE \(\downarrow\)\\ \end{tabular} & \begin{tabular}[c]{@{}l@{}}RMSE \(\downarrow\)\\ \end{tabular} & \begin{tabular}[c]{@{}l@{}}SD \(\downarrow\)\\ \end{tabular} & r \(\uparrow\) \\ \hline
CHROM~\cite{de_haan_robust_2013}       &     11.4                                            &         16.9                                             &              15.1                                      & 0.28  \\
POS~\cite{wang_algorithmic_2016}          &       11.5                                          &           17.2                                           &                15.3                                    &  0.30 \\
\hline
DeepPhys~\cite{chen_deepphys_2018}         &           11.0                                      &           13.8                                           &                 13.6                                   &  0.11 \\
PhysNet~\cite{yu2019remote}         &                 10.8                                &                 14.8                                     &                     14.9                               & 0.20   \\
RhythmNet~\cite{niu_rhythmnet_2020}    &                 5.30                                &                   8.14                                   &                     8.11                               &  0.76 \\
CVD~\cite{niu_video-based_2020}           &              5.02                                   &                    7.97                                  &                        7.92                            & 0.79   \\
Dual-GAN~\cite{lu_dual-gan_2021}      &                  \underline{4.93}                               &                  7.68                                    &                 7.63                                   &   \textbf{0.81}\\
BVPNet~\cite{das_bvpnet_nodate}        &                    5.34                             &                7.85                                      &        7.75                                            &  0.70 \\
\hline
Physformer~\cite{yu_physformer_2022}        &             4.97                                    &           7.79                                           &      7.74                                              & 0.78   \\
TransPPG \cite{kang_transppg_2022}       &             4.94                                   &           \underline{7.42}                                           &       \underline{7.44}                                             & 0.79   \\ 
\textbf{PhySU-Net (Ours)}   &  \textbf{4.53}                                                   &            \textbf{7.35}                                          &      \textbf{5.79}                                              & \underline{0.80} \\ \hline
\end{tabular}\label{tab1}
\end{center}
\end{table}

\textbf{Supervised}: We evaluate the supervised part of our method on the OBF and VIPL-HR datasets, as shown in Table~\ref{tab1}. We compare PhySU-Net to a wide array of previous supervised methods including traditional, CNN based and transformer based, showing that it reaches superior performance with RMSE of 0.659 on OBF and 7.35 on VIPL-HR. Our method's long-range temporal modelling proves effective on extracting an accurate HR on the challenging VIPL-HR data, with more reliable predictions than other methods with notably lower SD.

\begin{table}[t!]
\begin{center}
\setlength{\tabcolsep}{3pt}
\caption{Linear classification and Transfer learning on VIPL-HR~\cite{niu_vipl-hr_2018}}
\begin{tabular}{llllll}
\hline
Methods         & \begin{tabular}[c]{@{}l@{}}Data \\ Pretrain \hfill\(\rightarrow\)\hfill Train\end{tabular} & MAE \(\downarrow\)  & RMSE \(\downarrow\) & SD \(\downarrow\)  & r \(\uparrow\)    \\ \hline
Purely Supervised      & NONE\hfill \(\rightarrow\) VIPL                                                      & 4.53 & 7.35 & 5.79 & 0.80 \\ \hline
Linear classification & OBF \hfill \(\rightarrow\)\hfill VIPL                                                       & 6.19 & 9.16 & 6.75 & 0.68 \\ 
    (with Self-Supervision)            & VIPL \hfill\(\rightarrow\)\hfill VIPL                                                      & 6.30 & 9.28 & 6.81 & 0.67 \\ \hline
\begin{tabular}[c]{@{}l@{}}Transfer learning\\ (with Self-Supervision)\end{tabular}  & OBF \hfill\(\rightarrow\) \hfill VIPL                                                       & 4.22 & 7.07 & 5.66 & 0.82 \\ \hline
\end{tabular}
\label{tab2}
\end{center}
\end{table}

\noindent \textbf{Linear classification and transfer learning}: 
In Table~\ref{tab2}, with transfer learning we obtain a notable performance increase with RMSE reduced from 7.35 to 7.07, proving the transferable ability of our method, as the model learns useful representation on unlabeled OBF data that improve its performance when fine-tuning on VIPL-HR. 
Furthermore, in the linear classification the representations learned from the pretext task are of good quality as the performance is still satisfactory. Despite having a controlled environment, the representations learned on OBF's data yield 9.16 RMSE on VIPL-HR. Additionally, representations learned without labels on the challenging VIPL-HR data yield 9.28 RMSE.

\begin{table}[t!]
\begin{center}
\caption{Network design ablation on VIPL-HR~\cite{niu_vipl-hr_2018}.}
\begin{tabular}{lllll}
\hline
   & \begin{tabular}[c]{@{}l@{}}MAE \(\downarrow\) \\ \end{tabular} & \begin{tabular}[c]{@{}l@{}}RMSE \(\downarrow\)\\ \end{tabular} & \begin{tabular}[c]{@{}l@{}}SD \(\downarrow\)\\ \end{tabular} & r \(\uparrow\) \\ \hline
\textbf{Proposed (T = 576)} &                  \textbf{4.53}                                   &    \textbf{7.35}                                                  &        \textbf{5.79}                                            &  \textbf{0.80} \\
w/o HR head &             4.72                                       &     7.85                                                 &          6.28                                          &   0.78 \\
w/o Decoder  &            5.20                                         &        7.93                                              &       5.98                                             & 0.77   \\
w/o Stacking  &            5.62                                         &        8.35                                              &       6.17                                             & 0.74  \\
T = 384 &             4.80                                        &     7.80                                                 &          6.15                                          &   0.78 \\
T = 256 &            5.05                                         &        8.13                                              &       6.37                                             & 0.76   \\ \hline
\end{tabular}\label{tab3}
\end{center}
\end{table}

\begin{table}[t!]
\begin{center}
\caption{Transfer learning ablation study on VIPL-HR~\cite{niu_vipl-hr_2018}, with different self-supervised pre-training tasks on OBF~\cite{li_obf_2018}. Base method is underlined.}
\begin{tabular}{lllll}
\hline
\begin{tabular}[c]{@{}l@{}}Tasks:\\ Regression - Image\end{tabular} & \begin{tabular}[c]{@{}l@{}}MAE \(\downarrow\) \\ \end{tabular} & \begin{tabular}[c]{@{}l@{}}RMSE \(\downarrow\) \\ \end{tabular} & \begin{tabular}[c]{@{}l@{}}SD \(\downarrow\) \\ \end{tabular} & r \(\uparrow\)\\ \hline
No task  - No task      &     4.53                                       &                 7.35                              &  5.79                                               &         0.80                                           \\

CHROM  - No task      &     4.17                                       &                 7.17                              &  5.83                                                &         0.82                                           \\
No task  - Mask      &        4.46   & 7.32 & 5.80 & 0.81                                                       \\
\underline{CHROM  - Mask}            &       \underline{4.22}                                        &                 \underline{7.07}                                   &  \underline{5.66}                                                    &      \underline{0.82}                                               \\
GREEN  - Mask                                                   &    4.15                                                &        7.06                                              &     5.71                                               &  0.82 \\
LGI    - Mask                                                     &    4.17                                                &       7.08                                               &        5.72                                            &  0.82 \\
CHROM   - PBVP          &      4.35                                     &              7.18                                       &   5.71                                                  &          0.82                                             \\ \hline
\end{tabular}\label{tab4}
\end{center}
\end{table}

\noindent \textbf{Network components ablation:} In Table~\ref{tab3} rows 2 and 3, we show that both multitask learning components contribute to our method's effectiveness, as the HR head provides rough global supervision and the decoder performs fine-grained supervision on the signals. In row 4 of Table~\ref{tab2}, we show that performance worsens without stacking, meaning that it encourages the network to learn more informative features as attention windows also include signals that are further away in time, better exploiting the long-temporal context. \\
\noindent \textbf{Input sequence length ablation:} In Table~\ref{tab3} rows 5 and 6 we show that reducing the temporal context length leads to lower performance. The best performance is obtained with the longest T=576, showing that PhySU-Net is proficient at modeling long-temporal context. \\
\noindent \textbf{Transfer learning ablation and generalization:}
As shown in rows 2, 3 and 4 of Table~\ref{tab4} both pretext tasks contribute to a better downstream prediction as there is a drop in performance when excluding either of them.
Additionally, our method is generalizable as the pretext tasks can be easily changed. For the HR regression using different traditional methods (CHROM~\cite{de_haan_robust_2013}, LGI~\cite{pilz_local_2018}, GREEN~\cite{verkruysse_remote_2008}) yields similar results, as can be seen in Table~\ref{tab4} rows 4, 5 and 6. We also implement another image-based pretext task, in alternative to masking, called PBVPmap prediction. This task consists of predicting a PBVPmap, constructed from pseudo-BVP signals obtained with CHROM method. In row 7 of Table~\ref{tab4} we see that the PBVP task is also valid for pre-training, but masking yields stronger representations. Our self-supervised pre-training approach can be used with any kind of regression target and image-based task.

\section{Conclusion}
We propose PhySU-Net, a robust rPPG method that deals with challenging data by making full use of a long temporal context via our transformer model. With our proposed self-supervised pre-training framework, we further improve performance by leveraging unlabeled data. Experiments show that our supervised method is superior to other state-of-the-art methods. Moreover, with our generalized self-supervised pre-training framework, the model can learn meaningful representations from unlabeled data that are transferable. Future work can include the addition of new pretext tasks to our framework and usage of non-rPPG unlabeled data.

\section{Acknowledgement}
This work was supported by the Research Council of Finland (former Academy of Finland) for Academy Professor project EmotionAI (grants 336116, 345122) and ICT 2023 project TrustFace (grant 345948), and the University of Oulu \& Research Council of Finland Profi 7 (grant 352788). As well, the authors wish to acknowledge CSC – IT Center for Science, Finland, for computational resources.

{\small
\bibliographystyle{ieee}
\bibliography{egbib}
}

\end{document}